\documentclass[10pt,twocolumn,letterpaper]{article}

\usepackage{wacv}
\makeatletter
\@namedef{ver@everyshi.sty}{}
\makeatother
\usepackage{tikz}

\usepackage{times}
\usepackage{epsfig}
\usepackage{graphicx}
\usepackage{amsmath}
\usepackage{amssymb}
\usepackage{booktabs}

\usepackage[accsupp]{axessibility}  

\usepackage{epsfig}
\usepackage{wrapfig}
\usepackage{moreverb,url}
\usepackage{amsfonts,mathtools,leftidx}
\usepackage{multirow}
\usepackage{multicol}
\usepackage{mathrsfs}
\usepackage{nccmath} 
\usepackage{algorithm,algpseudocode}
\usepackage{xcolor,colortbl}
\usepackage{enumitem}
\usepackage[textwidth=10mm]{todonotes}
\usepackage{tabularx}
\usepackage[noadjust]{cite}
\usepackage{booktabs}

\def\wacvPaperID{793} 

\wacvalgorithmstrack   

\wacvfinalcopy 

\ifwacvfinal
\usepackage[breaklinks=true,bookmarks=false]{hyperref}
\else
\usepackage[pagebackref=true,breaklinks=true,colorlinks,bookmarks=false]{hyperref}
\fi

\begin{document}

\title{Centroid Distance Keypoint Detector for Colored Point Clouds}

\author{Hanzhe Teng, Dimitrios Chatziparaschis, Xinyue Kan, Amit K. Roy-Chowdhury, Konstantinos Karydis \\
University of California, Riverside\\
United States of America \\
{\tt\small \{hteng007, dchat013, xkan001, amitrc, karydis\}@ucr.edu}
\thanks{We gratefully acknowledge the support of NSF grants \#IIS-1724341 and \#IIS-1901379, ONR grant \#N00014-18-1-2252, and USDA/NIFA \#2021-67022-33453. Any opinions, findings, and conclusions or recommendations expressed in this material are those of the authors and do not necessarily reflect the views of the funding agencies.}
}

\maketitle
\thispagestyle{empty}

\begin{abstract}
Keypoint detection serves as the basis for many computer vision and robotics applications. Despite the fact that colored point clouds can be readily obtained, most existing keypoint detectors extract only geometry-salient keypoints, which can impede the overall performance of systems that intend to (or have the potential to) leverage color information. To promote advances in such systems, we propose an efficient multi-modal keypoint detector that can extract both geometry-salient and color-salient keypoints in colored point clouds. 
The proposed CEntroid Distance (CED) keypoint detector comprises an intuitive and effective saliency measure, the centroid distance, that can be used in both 3D space and color space, and a multi-modal non-maximum suppression algorithm that can select keypoints with high saliency in two or more modalities. The proposed saliency measure leverages directly the distribution of points in a local neighborhood and does not require normal estimation or eigenvalue decomposition. 
We evaluate the proposed method in terms of repeatability and computational efficiency (i.e. running time) against state-of-the-art keypoint detectors on both synthetic and real-world datasets. Results demonstrate that our proposed CED keypoint detector requires minimal computational time while attaining high repeatability. To showcase one of the potential applications of the proposed method, we further investigate the task of colored point cloud registration. 
Results suggest that our proposed CED detector outperforms state-of-the-art handcrafted and learning-based keypoint detectors in the evaluated scenes. 
The C++ implementation of the proposed method is made publicly available at 
\url{https://github.com/UCR-Robotics/CED_Detector}.
\end{abstract}

\section{Introduction}
Keypoint detection serves as the basis across computer vision and robotics applications such as 3D reconstruction~\cite{Henry2012mapping, Beall2010recon}, localization and mapping~\cite{Endres2014rgbdslam, Huang2016rgbd}, and navigation on point clouds~\cite{Gao2018flying, Krusi2017driving}. 
In such applications, colored point clouds can be obtained from RGB-D cameras, where color and depth images are aligned, or camera-LiDAR systems, where they are collected after extrinsic calibration; this process is often referred to as point cloud colorization or photometric reconstruction (e.g.,~\cite{Mirzaei2012calibration}).
However, most existing keypoint detectors only consider geometric information and fail to extract color-rich keypoints, which impedes the overall performance of systems that intend to (or have the potential to) leverage color information.
Hence, there is need for efficient keypoint detectors that can extract both geometric and color information in the point cloud. 

Keypoint detection aims to extract a subset of points from the point cloud so that they can best represent the original data in a compact form. 
Some successful keypoint detectors for 2D images, such as SIFT~\cite{kp_Lowe2004sift} and Harris~\cite{kp_Harris1988}, have been extended to 3D space following the original design ideas by the Point Cloud Library (PCL) community~\cite{Rusu2011pcl}.
However, the data structures used in 2D images and 3D point clouds are fundamentally different. 
This may limit deployment of such methods and has led to studies that focus on intrinsic properties of point clouds. 
Recent advances in deep learning has helped introduce several learning-based keypoint detectors and feature descriptors~\cite{kp_Li2019usip, kp_Yew2018featnet, fd_Zeng2017_3dmatch, Deng2018ppf-foldnet}.
In spite of their strong performance within their training domains, learned detectors and features may not transfer over in new scenes that are different from those used in training. For example, it may be challenging for a system trained with data collected in indoor environments to operate in diverse outdoor scenes~\cite{Cadena2016slam_survey}.

In contrast, methods that leverage the inherent properties of point clouds may help overcome this difficulty. 
Several existing methods focusing on geometric properties of point clouds, such as NARF~\cite{kp_Steder2011narf} and ISS~\cite{kp_Zhong2009iss}, require eigenvalue decomposition and/or normal estimation. 
These operations are computationally expensive, especially when performing keypoint detection at a large scale. 
In a distinct line of research, it has been shown that incorporating color modality can improve accuracy for applications such as point cloud registration~\cite{Park2017coloricp}. 

The main hypothesis underlying this work is that incorporating color modality (in addition to a geometric modality) can help improve the overall performance, as the amount of information passed on to the following components in the system has increased.
Despite existing descriptors that can incorporate color information (e.g.,~\cite{fd_Tombari2011shotcolor}), to the best of the authors' knowledge there currently exists no effective keypoint detector that can extract color-rich keypoints to feed to the descriptor. For instance, geometric-based keypoint detectors can fail to extract keypoints on a flat surface with color texture. 
While some methods (e.g., SIFT-3D) can extract color-rich keypoints, they do so at expense of losing geometric information (i.e. they can only respond to one modality). This limitation can be linked to lack of an effective non-maximum suppression algorithm to combine the two modalities; this is one of the key contributions of this work. 

To this end, we propose an efficient multi-modal keypoint detector, named CEntroid Distance (CED) keypoint detector, that utilizes both geometric and photometric information. 
The proposed CED detector comprises an intuitive and effective saliency measure, the centroid distance, that can be used in both 3D space and color space, and a multi-modal non-maximum suppression algorithm that can select keypoints with high saliency in two or more modalities. The proposed saliency measure leverages directly the distribution of points in a local neighborhood and does not require normal estimation or eigenvalue decomposition.
The proposed CED detector is evaluated in terms of repeatability and computational efficiency (running time) against state-of-the-art keypoint detectors on both synthetic and real-world datasets. Results demonstrate that our proposed CED keypoint detector requires minimal computational time while attaining high repeatability. 
In addition, to showcase one of the potential applications of the proposed method, we further investigate the task of colored point cloud registration. 
Results show that our CED detector outperforms state-of-the-art crafted and learning-based keypoint detectors in the evaluated scenes.

\noindent\textbf{Contributions.} The paper's contributions are fourfold:
\begin{itemize}
\item We propose an efficient multi-modal keypoint detector that can extract both geometry-salient and color-salient keypoints in a colored point cloud, with the potential to be extended and applied to point clouds with multiple modalities (e.g., colored by multi-spectrum images). 
\item We propose to use an intuitive and effective measure for keypoint saliency, the distance to centroid, which can leverage directly the distribution of points and does not require normal estimation or eigenvalue decomposition.
\item We develop a multi-modal non-maximum suppression algorithm that can select keypoints with high saliency in two or more modalities.
\item We demonstrate through experiments in four datasets that the proposed keypoint detector can outperform the state-of-the-art handcrafted and learning-based keypoint detectors.
\end{itemize}

\section{Related Work}
3D keypoint detectors can be categorized as those extending designs originally developed for 2D images~\cite{kp_Harris1988, kp_Lowe2004sift}, and those native to 3D point clouds~\cite{kp_Chen2007LSP, kp_Zhong2009iss, kp_Mian2010KPQ, kp_Sun2009HKS} and 3D meshes~\cite{kp_Unnikrishnan2008LBSS, kp_Zaharescu2009MeshDoG, kp_Castellani2008SP}.
Following the design in 2D images, Harris family~\cite{kp_Harris1988} computes covariance matrices of surface normal or intensity gradient in 3D space, and in 3D and color space (herein referred to as 6D space). SIFT~\cite{kp_Lowe2004sift} applies the difference of Gaussian operator in scale-space to find keypoints with local maximal response. 
However, for 3D point clouds, the amount and position of points within the spherical region are uncertain, making it hard to obtain gradients. 
In 3D space, Normal Aligned Radial Feature (NARF)~\cite{kp_Steder2011narf} measures saliency from surface normal and distance changes between neighboring points. Intrinsic Shape Signature (ISS)~\cite{kp_Zhong2009iss} and KeyPoint Quality (KPQ)~\cite{kp_Mian2010KPQ} perform eigenvalue decomposition of the scatter matrix of neighbor points and threshold on the ratio between eigenvalues.
Heat Kernel Signature (HKS)~\cite{kp_Sun2009HKS} and  Laplace-Beltrami Scale-space (LBSS)~\cite{kp_Unnikrishnan2008LBSS} measure the saliency from the response to the Laplace-Beltrami operator in the neighborhood.
Local Surface Patches (LSP)~\cite{kp_Chen2007LSP} leverages local principal curvatures to construct the Shape Index (SI)~\cite{kp_Dorai1997SI_ShapeIndex} as the measure of saliency. 
As in SIFT, MeshDoG~\cite{kp_Zaharescu2009MeshDoG} and Salient Points (SP)~\cite{kp_Castellani2008SP} apply the difference-of-Gaussian operator to construct the scale space for saliency measure. 
We refer readers to the comprehensive evaluation in~\cite{kp_Tombari2013survey_ijcv} for more details. 

In summary, the existing methods often apply an operator to obtain point normal, curvature and gradient in the local region, and threshold on either a combination of the obtained measures or the eigenvalues of the covariance matrices. 
On the contrary, our proposed method leverages directly the point distribution and statistics in 3D space and color space, without the need of normal estimation or eigenvalue decomposition.

Learning-based approaches, such as USIP~\cite{kp_Li2019usip} and 3DFeat-Net~\cite{kp_Yew2018featnet}, have also been studied. 3DFeat-Net~\cite{kp_Yew2018featnet} learns a 3D feature detector and descriptor for point cloud matching using weak supervision, whereas USIP~\cite{kp_Li2019usip} trains a feature proposal network with probabilistic Chamfer loss in an unsupervised manner. 

Due to the constraint on quantization in neural networks, the extracted keypoints may be non-deterministic given the same point cloud input (e.g., USIP~\cite{kp_Li2019usip}). 
Despite the relaxation of the requirement of ground truth, generalization capability may be challenged when deploying the system in practice. 
These approaches are often trained with data collected from target environments and/or designated sensors, and the performance with cross-source data in unseen scenes is unclear and less reported~\cite{Huang2021reg_survey}.
In addition, the high computational requirements in deep networks necessitate on-board GPUs for real-time applications, which is challenging for small robots/systems~\cite{Mohta2017uav,Gao2018flying} with limited computational resources (constrained by payload and battery life). 
To facilitate the deployment in various scenes and enable real-time execution, our proposed method is designed to focus on the invariant features in point clouds while keeping the computational time to a minimum.

Color modality has been taken into account in descriptors~\cite{fd_Tombari2011shotcolor} as well as in registration algorithms~\cite{Park2017coloricp, Huhle2008colorndt}. Even though color modality offers promising prospects in the aforementioned tasks, it is rarely incorporated into the design of keypoint detectors, and in turn impedes the effectiveness of methods building on top of keypoint detectors. Existing 3D keypoint detectors with color modality (e.g., Harris-6D) combine three-channel RGB responses into intensity by a weighting function; doing so leads to loss of information and hence such methods cannot capture color variations effectively. One potential reason behind this is that the commonly used non-maximum suppression algorithm can only threshold on one modality, and therefore all factors to be considered have to be weighted together into a unitary function before thresholding. In our proposed method, we consider the saliency measure in 3D space and color space separately, and then select the best response by a multi-modal non-maximum suppression algorithm that can threshold on two or more unitary functions at the same time.

\section{Centroid Distance (CED) Keypoint Detector}

In this section we describe our proposed CEntroid Distance (CED) keypoint detector for colored point clouds based on an efficient measure of keypoint saliency--the distance to centroid. We discuss the measure in 3D space (Section~\ref{sec_ced_3d}) and RGB color space (Section~\ref{sec_ced_color}). Then, the two saliency measures are combined via a multi-modal non-maximum suppression algorithm (Section~\ref{sec_non_max}) to select the best local keypoint. 

Consider a point $\mathbf{p}=[\mathbf{{}^g p},  \mathbf{{}^c p}]^\mathrm{T}$ that consists of a geometric component $\mathbf{{}^g p} = \{p_x, p_y, p_z\}$, and a color component $\mathbf{{}^c p} = \{p_r, p_g, p_b\}$, where $p_x, p_y, p_z \in \mathbb{R}$ and $p_r, p_g, p_b \in [0, 1]$. A colored point cloud is defined as $\mathbf{P} = \{\mathbf{p_1}, \mathbf{p_2}, \ldots, \mathbf{p_n} \}$. Recall that the task of keypoint detection is to extract a subset from $\mathbf{P}$ such that the extracted keypoints can potentially best represent the original data in a compact yet effective form. The typical pipeline for keypoint detection is to examine every point in $\mathbf{P}$ to obtain a measure of saliency (e.g., cornerness) by evaluating its relation with neighbors in the support (i.e. search radius). Then, a non-maximum suppression algorithm is used to disable keypoints that are not salient compared with their neighbors, and at the same time, keep those with maximum values of saliency. Following this pipeline, we introduce each component of the proposed CED keypoint detector.

\begin{figure}[t]
\vspace{6pt}
\centering
\includegraphics[width=0.9\linewidth]{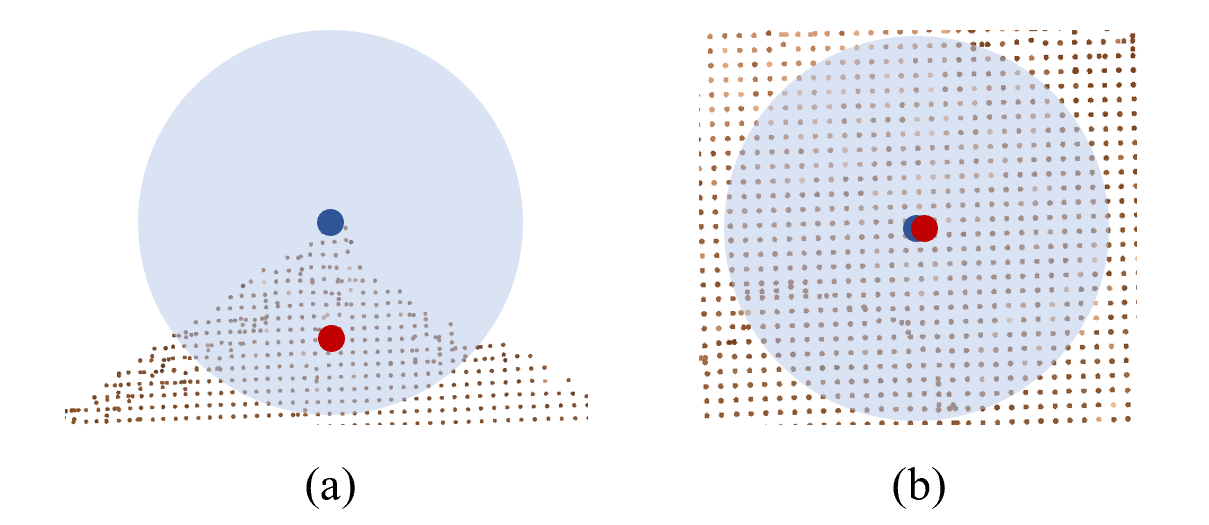}
\caption{Query point (blue) and geometric centroid (red) of all points in the support for (a) a query point located at corner, and (b) a query point located in the middle of a flat surface. 
Points in background (brown) are extracted from real-world datasets. 
}\label{fig_centroid_3d}
\end{figure}

\subsection{Centroid Distance in 3D Space}\label{sec_ced_3d}
We first consider the geometric component of a point $\textbf{p}$. Corners are among the most commonly used features in vision-based tasks. One notable measure of the orientation of a corner in 2D images is the intensity centroid~\cite{kp_Rosin1999centroid}. In this work, we consider the centroid in 3D space, as well as color space (more details in Section~\ref{sec_ced_color}). Different from those methods that take the intensity centroid as the measure of orientation, e.g., ORB~\cite{kp_Rublee2011orb}, we observe that the distance from the origin to the centroid can be a stable measure of cornerness in 3D space.

Given a point cloud $\mathbf{P}$, and a query point $\mathbf{p_i} \in \mathbf{P}$, the set of neighbor points $\mathbf{p_j}$ within a search radius $r$ near the query point $\mathbf{p_i}$ is defined as $N_g = \{ \mathbf{p_j} \mid \left\lVert \mathbf{{}^g p_i} - \mathbf{{}^g p_j} \right\rVert_2 < r\}$. The geometric centroid of the spherical region $\mathbf{{}^g} \boldsymbol{\mu_{p_i}}$ is
\begin{equation}\label{eq_centroid}
\mathbf{{}^g} \boldsymbol{\mu_{p_i}}=\frac{1}{|N_g|} \sum_{N_g} \mathbf{{}^g p_j}\enspace,
\end{equation} 
where $|N_g|$ is the cardinality of set $N_g$. The geometric centroid of a spherical region is equivalent to the mean over all points in $N_g$ in each dimension of $\mathbf{{}^g p_j}$. 

We compute the distance between the query point, $\mathbf{p_i}$, and the geometric centroid of its neighbors, $\mathbf{{}^g}\boldsymbol{\mu_{p_i}}$, using the L2 norm, i.e. $d_g=\left\lVert\mathbf{{}^g p_i} - \mathbf{{}^g} \boldsymbol{\mu_{p_i}}\right\rVert_2$. Figure~\ref{fig_centroid_3d}(a) demonstrates an example of a corner, where the query point (blue) has the maximum response in the vicinity, i.e. the largest distance to the geometric centroid (red). On the other hand, Figure~\ref{fig_centroid_3d}(b) shows a flat surface, where the geometric centroid (red) is in close proximity to the query point (blue). It is evident that the greater this distance is, the more salient it is as a corner point. 

\textbf{Geometric Centroid Invariance.}
As the relative position (distance and angle) of all points with respect to each other within a sphere is fixed regardless of the translation and rotation of the sphere, the relative position of the geometric centroid of all points is also fixed with respect to the center of sphere, at which the query point is located. Hence, the L2 norm from the query point to its geometric centroid, used as the saliency measure in 3D space, is invariant to rotation and translation.

\begin{figure}[t]
\centering
\includegraphics[width=0.9\linewidth]{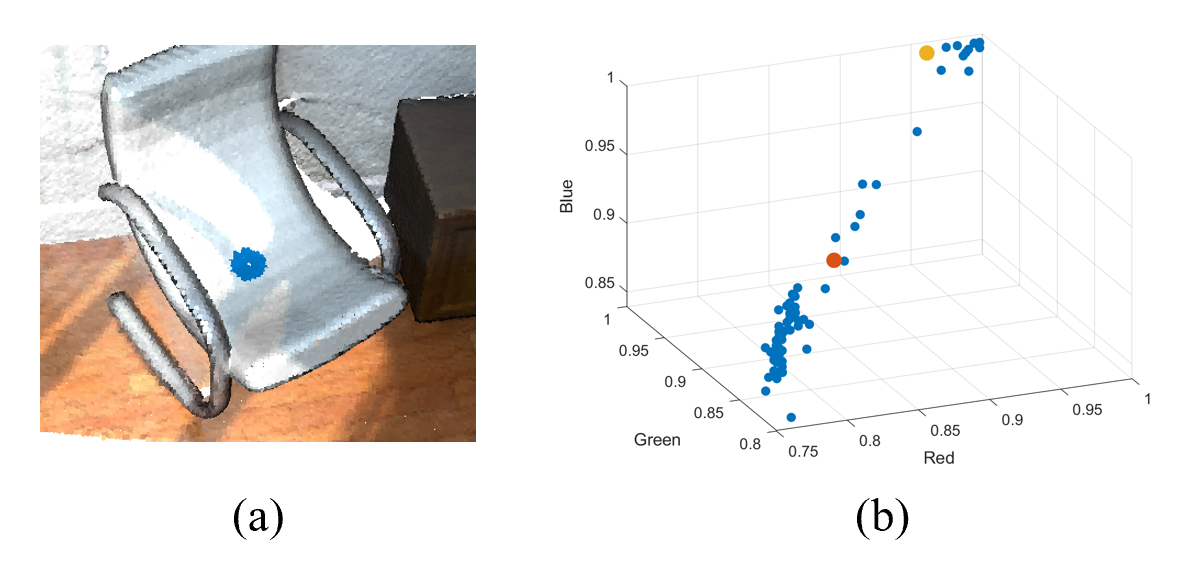}
\caption{(a) A detected keypoint located at position with significant color variation (spherical region colored in blue). (b) RGB color space graph for neighbor points (blue) in the support of the query point (yellow) and the photometric centroid (red) according to the detected keypoint in (a).}
\label{fig_centroid_color}
\end{figure}

\subsection{Centroid Distance in Color Space} \label{sec_ced_color}
Besides the geometric component, we can also consider the color component in colored point clouds (when available) by extending the concept of centroid to color space as a separate measure. 
Given a query point $\mathbf{p_i}$, we first identify its neighbor points $\mathbf{p_j}$ as discussed in Section~\ref{sec_ced_3d}. 
Then, we collect the color components of selected neighbor points (i.e. $\mathbf{^c p_j}$, $j=1,\ldots,|N_g|$) into a three-dimensional RGB color space set. The goal is to find the points with greatest color variation within this set.
Similar to~\eqref{eq_centroid}, we can then compute the photometric centroid $\boldsymbol{\mathbf{{}^c} \mu_{p_i}}$ as
%
\begin{equation}\label{eq_color_centroid}
\boldsymbol{\mathbf{{}^c} \mu_{p_i}}=\frac{1}{|N_g|} \sum_{N_g} \mathbf{{}^c p_j}\enspace.
\end{equation} 
We consider the L1 norm for the photometric centroid distance $d_c=\left\lVert\mathbf{{}^c p_i} - \boldsymbol{\mathbf{{}^c} \mu_{p_i}}\right\rVert_1$
to measure if a point has significant color changes within its neighborhood. 
\footnote{We acknowledge that a perceptually uniform space (e.g., CIELAB) may be a better choice when measuring distance in color space. However, measuring directly in RGB space is more efficient, and the performance difference is negligible when color variation is significant.}

Figure~\ref{fig_centroid_color}(a) illustrates an example of a detected keypoint with neighboring points of diverse color properties. The corresponding RGB color space graph of all points in the spherical region is depicted in Figure~\ref{fig_centroid_color}(b).
Here, the photometric centroid (red) is closer to the majority of points sharing similar color. The query point (yellow) is located in the spare side, where points have larger point-to-centroid distance and are highly likely to be chosen in the following non-maximum suppression step in this neighborhood.

\textbf{Photometric Centroid Invariance.} 
Regardless of rotation and translation in 3D space, in color space the distribution of all points within the sphere remains unchanged. Hence, the L1 distance from the query point to the photometric centroid employed in color space is invariant to rotation and translation, as the measure of saliency.

\begin{figure}[t]
\vspace{3pt}
\centering
\includegraphics[width=0.6\linewidth]{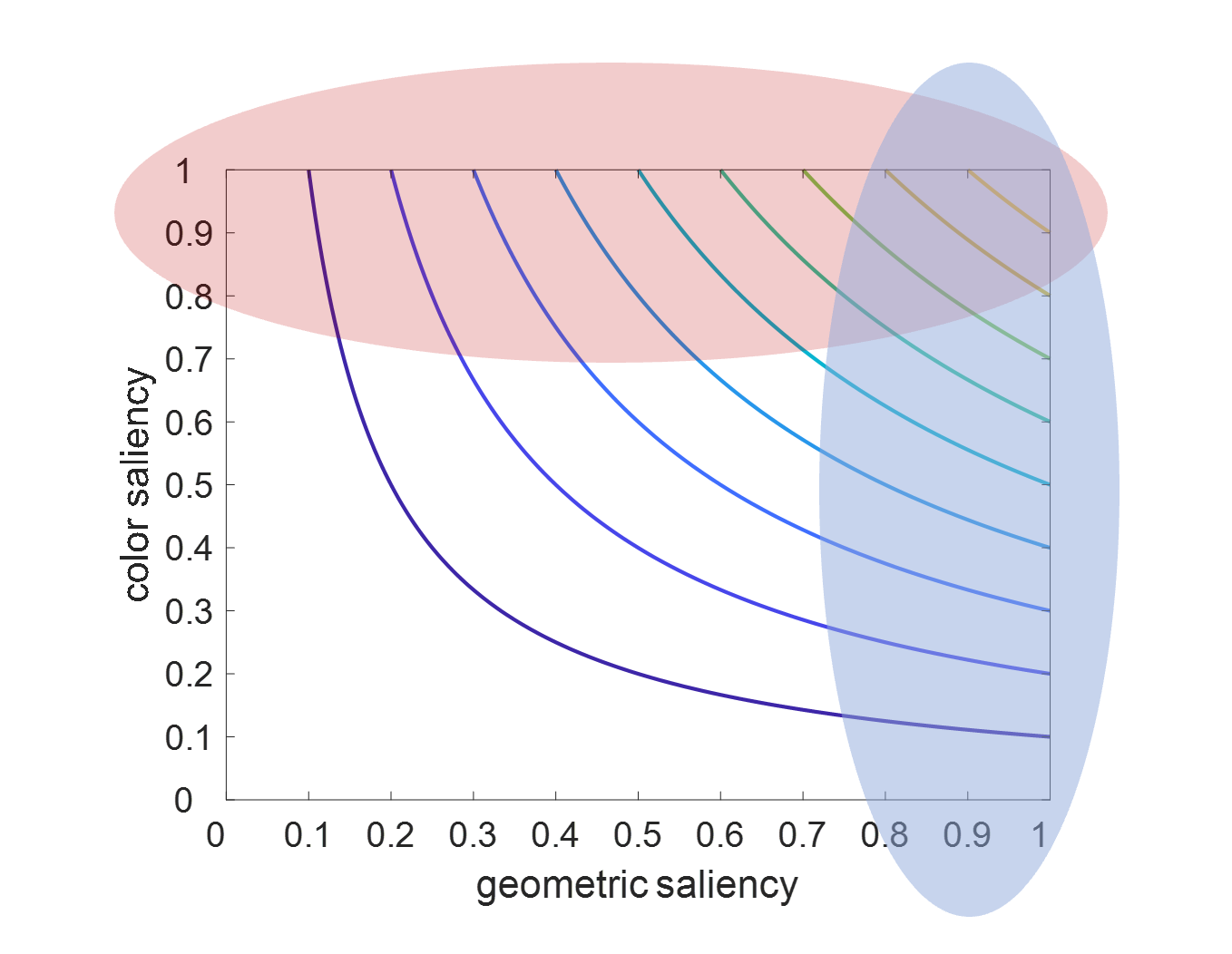}
\caption{The contour plot for the multiplication of two saliency measures. The light blue zone reflects the geometry-salient keypoints, whereas the light red zone reflects the color-salient keypoints.}
\label{fig_non_max}
\end{figure}

\begin{table*}[t]
\begin{center}
\begin{tabular}{ccccc}
\toprule
Dataset & Redwood Synthetic & Redwood Scan & TUM & SUN3D \\
\midrule
Type  & synthetic  & real & real & real \\
Scene & RGB-D frame  & stitched fragment & RGB-D frame & reconstruction \\
Ave. \# Points   & 101164 & 63841  & 63211 & 56497 \\
Resolution (m)   & 0.01   & 0.01 & 0.01 & 0.1 \\
Noise added (m)  & 0.005  & 0.005 & 0.005 & 0.05 \\
$\epsilon$ in Eq.~\ref{eq_rep} (m) & 0.02 & 0.02 & 0.02 & 0.2 \\
\bottomrule
\end{tabular}
\end{center}
\caption{Dataset Characteristics and Parameters}
\label{table_dataset}
\end{table*}

\subsection{Multi-modal Non-maximum Suppression}\label{sec_non_max}
Once the saliency measures in 3D space and color space are both obtained, the next step is to 
determine if the query point is a salient keypoint by utilizing and integrating the two measures. The standard approach for single-response cases is to check if the current response is the maximum among its neighbors. Here, we generalize that notion to multi-modal responses. Our approach is outlined in Algorithm~\ref{algorithm1}. The contour plot for the multiplication of two saliency measures is visualized in Figure~\ref{fig_non_max}. 

\begin{algorithm}[h!]
\caption{Multi-modal Non-maximum Suppression }\label{algorithm1}
\begin{algorithmic}[1]
\State \textbf{Input}: point cloud $\mathbf{P}$, saliency measure in 3D space $d_{gi}$ and in color space $d_{ci}$ for $\mathbf{p_i} \in \mathbf{P}$, threshold for 3D space $t_g$ and for color space $t_c$
\State \textbf{Output}: keypoints $\mathbf{K}$
\State \textbf{Initialization}: $\mathbf{K} \gets \{\}$
\For {each $\mathbf{p_i} \in \mathbf{P}$}
\If{$d_{gi} < t_g$ \textbf{and} $d_{ci} < t_c$} 
 \State  \textbf{continue}
\EndIf
\State maximum $\gets$ \textit{true}
\For {each $\mathbf{p_j} \in N_g$}
\If{$d_{gi} \cdot d_{ci} < d_{gj} \cdot d_{cj}$}
\State maximum $\gets$ \textit{false}
\EndIf
\EndFor
\If{maximum \textbf{is} \textit{true}}
\State $\mathbf{K} \gets \mathbf{K} \cup \{\mathbf{p_i}\}$
\EndIf
\EndFor
\end{algorithmic}
\end{algorithm}

Recall that the goal is to extract geometry-salient (blue ellipse in Figure~\ref{fig_non_max}) and/or color-salient (red ellipse in Figure~\ref{fig_non_max}) keypoints, which means we cannot simply remove those points that underperform in one modality, since points located at the center of a flat surface might contain rich texture and points with neighbors in same color might be located at corners. Therefore, lines 5-7 in Algorithm~\ref{algorithm1} are used to filter those points that are neither geometry-salient nor color-salient by the AND logic. After this step, points outside the red and blue ellipses, which might affect the following contour optimization, are discarded. 

The next step, lines 8-13 in Algorithm~\ref{algorithm1}, is to select the point that has the best performance considering both geometric and color modalities. This is achieved by optimizing the contour, constructed from the multiplication of the two saliency measures. 
The intuition is that in the case of a flat surface, for example, the saliency measures in 3D space are of the same magnitude, hence the points with higher color response will be selected as keypoints (multiplying with a same value does not change the order of color response). In the cases when the colors of all points are similar (e.g., table corner), the comparison becomes to mainly consider the geometric response. 

Though what we illustrate in Algorithm~\ref{algorithm1} in its current form is for two modalities, the algorithm by itself is designed for multiple modalities (e.g., in the case of multi-spectrum perception). The multi-modal form can be obtained by extending line 5 and line 10 in Algorithm~\ref{algorithm1} with more saliency measures.

\section{Evaluation and Results}
We present the qualitative results of the proposed method against others in Section~\ref{sec_qualitative}, followed by quantitative evaluation in terms of repeatability (Section~\ref{sec_repeatability}) and computational efficiency (Section~\ref{sec_computation}). 
To showcase one of the potential applications of the proposed CED detector, we present the results of colored point cloud registration in Section~\ref{sec_registration}. The ablation study of the proposed method is discussed in Section~\ref{sec_ablation}.

\subsection{Experiment Setup}
In experiments, we consider both CED and CED-3D, which is a variant of CED that considers only geometric information for non-colored point clouds. We evaluate the performance of CED and CED-3D detectors against ISS~\cite{kp_Zhong2009iss}, SIFT-3D~\cite{kp_Lowe2004sift}, Harris-3D and Harris-6D~\cite{kp_Harris1988} detectors
\footnote{The implementation of these methods is available in PCL 1.8, where SIFT-3D, Harris-3D and Harris-6D are implemented according to their original ideas in 2D space.} as well as a state-of-the-art learning-based keypoint detector USIP~\cite{kp_Li2019usip}.
\footnote{The source code and pre-trained network models are provided by the authors at https://github.com/lijx10/USIP.}
A random keypoint selector is also included as baseline for comparison.
Experiments are conducted with an i7 8th-gen CPU and Ubuntu 18.04 operating system. An additional Quadro P1000 GPU with CUDA 10.2 support is provided for the USIP detector only. The evaluation of computational efficiency is performed consistently using one thread.

We use four datasets, Redwood Synthetic~\cite{Choi2015recon_redwood}, Redwood Scan~\cite{Park2017coloricp}, TUM~\cite{Sturm2012dataset_tum} and SUN3D~\cite{Xiao2013sun3d} for evaluation. The selected datasets span various scenarios including different type, scale, environment, and number of points. During pre-processing, point clouds are downsampled to their minimum resolution and NaN points are removed. 
Features and key parameters for each dataset are shown in Table~\ref{table_dataset}. Parameters with respect to each method are tuned to their best performance or selected as recommended by the authors of the respective works. (We provide more details in the parameter tuning process for each method in the supplementary materials.)

\begin{figure*}[t]
\vspace{0pt}
\centering
\includegraphics[width=\linewidth]{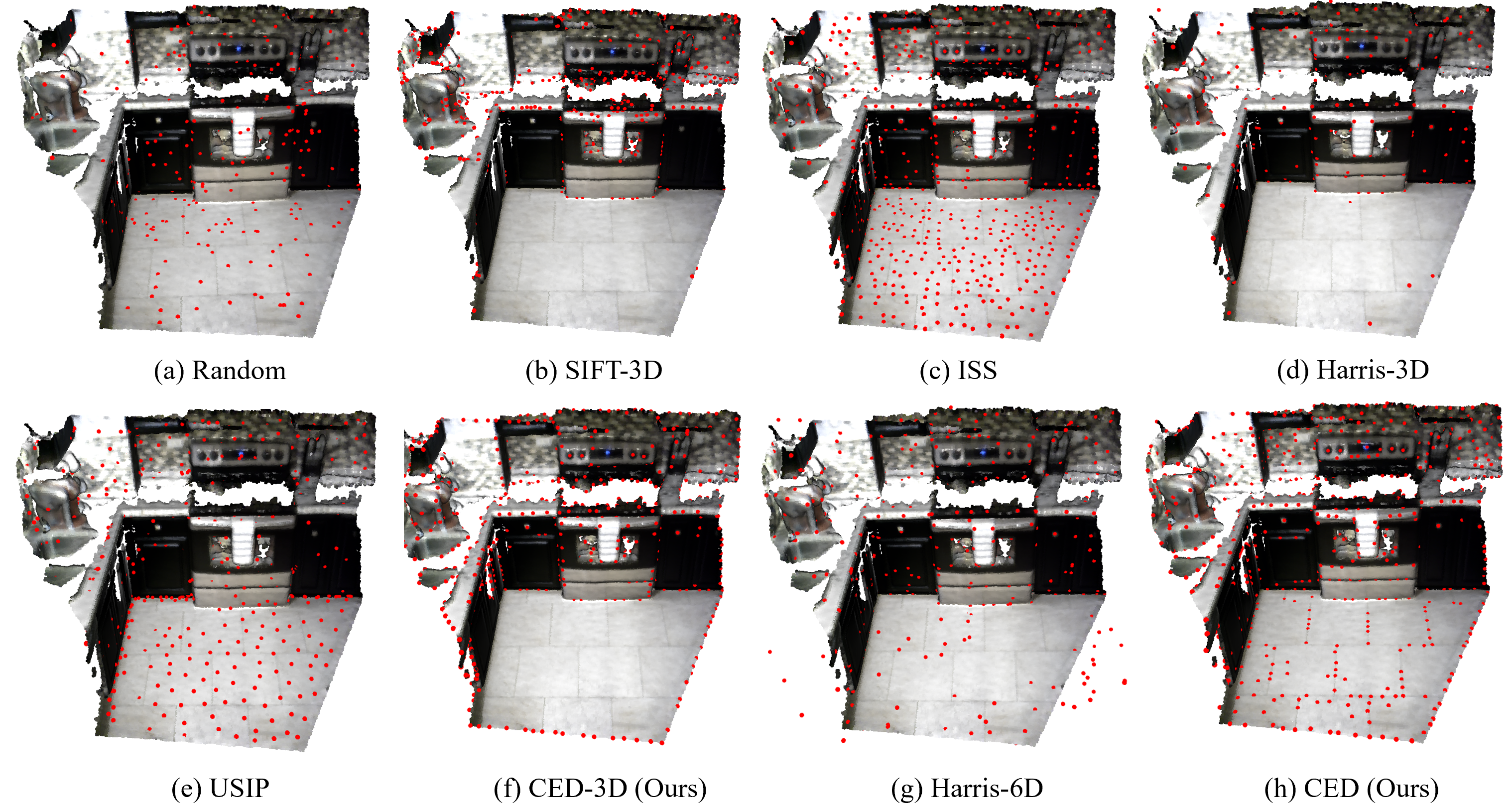}
\vspace{2pt}
\caption{Qualitative evaluation of the proposed CED and CED-3D (its geometry-only variant) keypoint detectors against other methods on Redwood Scan~\cite{Park2017coloricp} dataset. 
(a-f) Methods able to extract geometry-salient keypoints only. (g-h) Methods able to extract both geometry- and color-salient keypoints. 
Two key observations can be made in this comparison. 1) Out of all methods, only ISS, CED-3D and CED can extract keypoints on four stove knobs. 2) CED can capture \textbf{color changes between floor tiles} and extract keypoints with \textbf{high regularity}, whereas ISS and USIP extract keypoints in a somehow uniform manner, and other methods are not capable of extracting meaningful keypoints on the floor.}
\label{fig_qualitative}
\end{figure*}

\subsection{Qualitative Evaluation}\label{sec_qualitative}
We select an arbitrary frame in the Redwood Scan~\cite{Park2017coloricp} dataset for qualitative evaluation. (The supplementary materials contain qualitative analysis for a few more randomly-picked frames across datasets.)
Comparative results are shown in Figure~\ref{fig_qualitative}.
Three main observations can be made.
\begin{itemize}
\item  The random keypoint detector produces keypoints without using any geometric or color information in the point cloud. 
The USIP detector takes in non-colored point clouds and proposes candidate interest positions in 3D space (instead of selecting existing points on point clouds). 
The keypoints proposed by both detectors lack physical meaning and are non-deterministic given the same input cloud.

\item  SIFT-3D, ISS, Harris-3D and CED-3D make use of only geometric information, and can capture corners and edges in the scene. Note that only ISS and CED-3D can produce stable, regular keypoints at geometry-salient places such as the four stove knobs.
However, ISS produces many meaningless keypoints on the floor due to its sensitivity to the noise on planar surfaces.
As opposed to other methods, CED-3D provides meaningful keypoints with high regularity and is shown to be stable at planar surfaces.

\item  Harris-6D and CED can further leverage color information in the colored point cloud. Harris-6D computes color-salient keypoints using intensity gradient, but in a somehow random pattern (note that some Harris-6D keypoints are located out of the frame). In contrast, CED produces highly regular geometry- and color-salient keypoints using directly the distribution in 3D space and color space, and can clearly capture color changes between floor tiles. These regular keypoints can even serve as edge features in geometric tasks when needed. This is the key to improve the overall performance of systems that intend to leverage color information.
\end{itemize}

\begin{figure*}[t]
\vspace{-3pt}
\centering
\includegraphics[width=\linewidth]{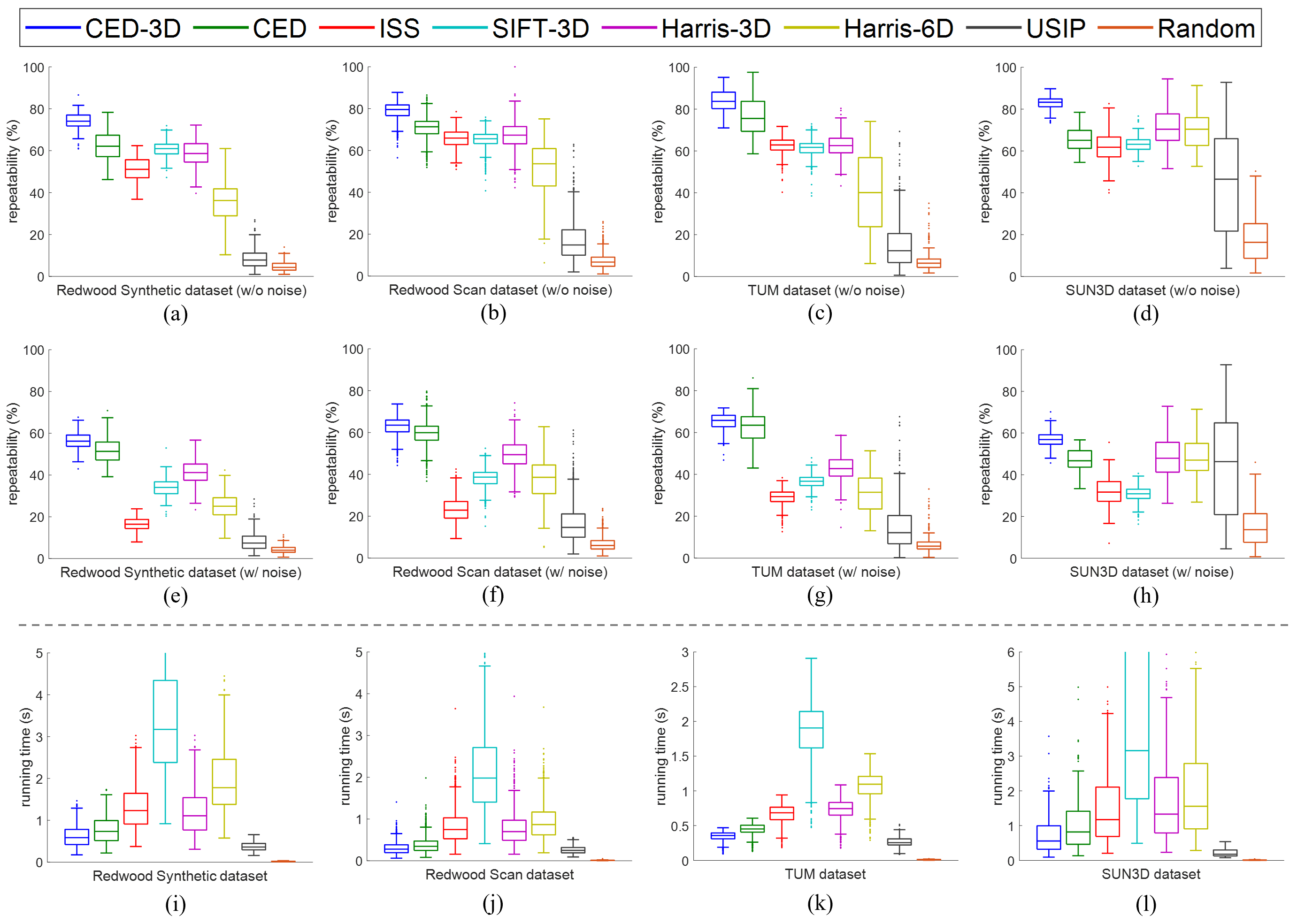}
\caption{The evaluation result of relative repeatability without noise added (a-d), with noise added (e-h) and running time (i-l) for Redwood Synthetic~\cite{Choi2015recon_redwood}, Redwood Scan~\cite{Park2017coloricp}, TUM~\cite{Sturm2012dataset_tum} and SUN3D~\cite{Xiao2013sun3d} datasets respectively.}
\label{fig_repeatability_runtime}
\end{figure*}

\subsection{Repeatability}\label{sec_repeatability}
Repeatability refers to the capability to extract same keypoints under various disturbances. A stable keypoint detector is expected to be invariant to translation, rotation and noise. 

We evaluate the repeatability of the proposed methods following~\cite{kp_Tombari2013survey_ijcv}. Given a point cloud $\mathbf{P}$, we obtain point cloud $\mathbf{Q}$ by applying an arbitrary transformation matrix $T \in SE(3)$ to $\mathbf{P}$. After this step, Gaussian noise can be added to $\mathbf{Q}$ if needed in the evaluation. We then extract keypoints in $\mathbf{P}$ and $\mathbf{Q}$ to obtain keypoint sets $\mathbf{K_P}$ and $\mathbf{K_Q}$ respectively.
A keypoint $\mathbf{p_i} \in \mathbf{K_P}$ is said to be \textit{repeatable} if the L2 distance between the geometric components of the transformed $\mathbf{p_i}$ and its nearest neighbor $\mathbf{q_j} \in \mathbf{K_Q}$ is less than a threshold $\epsilon$, i.e.
\begin{equation}\label{eq_rep}
    \left\lVert T\ \mathbf{{}^g p_i} - \mathbf{{}^g q_j} \right\rVert_2 < \epsilon\enspace.
\end{equation}
The relative repeatability then is the percentage of repeatable keypoints in $\mathbf{K_P}$. 

We follow the above steps to obtain the results of relative repeatability over four datasets. The standard deviation of Gaussian noise is set to be half of the resolution of point clouds,
and the repeatability threshold $\epsilon$ is set to be twice of the resolution of point clouds (see Table~\ref{table_dataset} for details). 
As shown in Figure~\ref{fig_repeatability_runtime}(a-h), CED and CED-3D outperform other methods by achieving the highest relative repeatability. In the case that Gaussian noise is added, our proposed methods are robust to noise, whereas significant performance degradation is observed for other methods. Additional testing for varying levels of noise intensity (up to twice the respective dataset resolution) is included in the supplementary materials, demonstrating that as the noise intensity begins exceeding the point cloud resolution, performance of all methods degrades as noise intensity increases. 

\subsection{Computational Efficiency}\label{sec_computation}
By computational efficiency we consider the running time required to extract keypoints in a point cloud. The temporal duration is recorded for each point cloud over all sequences in four datasets.

Results in Figure~\ref{fig_repeatability_runtime}(i-l) suggest that CED and CED-3D on average keep the computation to a minimum. The random keypoint detector achieves the fastest running time (as expected); however, it has very low repeatability close to zero (shown in Figure~\ref{fig_repeatability_runtime}(a-h)). 
Although the USIP keypoint detector requires similar or less time than CED and CED-3D, it requires a GPU for computation and its repeatability is significantly lower than that of CED and CED-3D.

\subsection{Point Cloud Registration}\label{sec_registration}
The proposed CED keypoint detector can be utilized in applications such as 3D reconstruction, SLAM and odometry. To showcase one of the potential applications, we present herein results of colored point cloud registration on the Redwood Synthetic~\cite{Choi2015recon_redwood} dataset.

We follow a typical registration pipeline consisting of the following four steps. 
1) For each pair of point clouds that describe the same scene from different view point, we apply the aforementioned eight keypoint detectors (in Sections~\ref{sec_repeatability} and~\ref{sec_computation}) to extract keypoints. 
\footnote{For USIP, the keypoints are selected as the closest points to its proposed candidate positions in 3D space.}
2) The feature descriptor PFHRGB~\cite{fd_Rusu2008pfh} (the color counterpart of the popular FPFH~\cite{fd_Rusu2009fpfh} descriptor) is used to describe the keypoints.
3) Correspondences are established by nearest neighbor search in feature vector space using \textit{k}-D tree.
4) Given the established correspondences, TEASER~\cite{Yang2020teaser} is used to estimate a transformation that can best align the two point clouds. Registration is successful if the translation error between the estimated transformation and the ground truth is less than a threshold of $0.2$\;m and the rotation error is less than a threshold of $5$\;deg; thresholds were selected based on values used in relevant literature, e.g.,~\cite{kp_Li2019usip}.

Results of registration success rates for the Redwood Synthetic dataset are shown in Table~\ref{table_reg}. We observe that CED outperforms other methods in all scenes, and CED-3D achieves competitive performance by using only the geometric information in the environments. 
The performance improvement indicates that our CED detector is capable of capturing a few more scenes where other methods fail to extract meaningful keypoints. 
Note that USIP proposes interest positions in 3D space as keypoints, as oppose to selecting points on the point cloud. This behavior is non-deterministic for the same point cloud and it can severely impede the performance of point cloud registration. Reported results are the best among all trained network models. (Additional details on the USIP's performance with sub-performing network models are provided in the supplementary materials.)

\begin{table}[h]
\begin{center}
{\small{
\begin{tabular}{ccccc}
\toprule
Method $\backslash$ Sequence 
& liv rm1 & liv rm2 & office1 & office2  \\
\midrule
Random    & 3.57  & 15.22 & 11.54 & 16.33 \\
SIFT-3D   & 32.14 & 23.91 & 57.69 & 59.18 \\
ISS       & 46.43 & 56.52 & 59.62 & 75.51 \\
Harris-3D & 67.86 & 78.26 & 80.77 & 81.63 \\
Harris-6D & 71.43 & 78.26 & 84.62 & 79.59 \\
USIP      & 37.50 & 52.17 & 57.69 & 73.47 \\
\midrule
CED-3D (Ours) & 60.71 & 56.52 & 80.77 & 79.59\\
CED (Ours)    & \textbf{76.79} & \textbf{80.43} & \textbf{86.54} & \textbf{83.67} \\
\bottomrule
\end{tabular}
}}
\end{center}
\caption{Success Rates (\%) of Colored Point Cloud Registration}
\label{table_reg}
\end{table}

\subsection{Ablation Study}\label{sec_ablation}
Our CED keypoint detector depends on two thresholding parameters, one for the distance to geometric centroid and one for the distance to photometric centroid. (See $t_g$ and $t_c$ at the line 5 in Algorithm~\ref{algorithm1}.) The geometric threshold ranges from 0 to 1, representing the ratio of the L2 distance to centroid over the radius of the spherical neighborhood region. The photometric threshold ranges from 0 to 3, representing the L1 distance variation in color space.

We evaluate on Redwood Synthetic~\cite{Choi2015recon_redwood} dataset for geometric threshold ranging from 0.1 to 0.5 when photometric threshold is fixed at 0.1 (Table~\ref{table_ablation_geometric}), and for photometric threshold ranging from 0.1 to 0.5 when geometric threshold is fixed at 0.2 (Table~\ref{table_ablation_color}). We observe that the number of keypoints extracted decreases as the threshold increase. The repeatability achieves the highest when $t_g = 0.2$ and $t_c = 0.5$. The running time is shown to be less dependent on the two thresholds.

\begin{table}[h]
\begin{center}
{\small{
\begin{tabular}{cccccc}
\toprule
$t_g$ & 0.1 & 0.2 & 0.3 & 0.4 & 0.5 \\
\midrule
\# Keypoints   & 669.70  & 616.20 & 584.18 & 529.32 & 488.65 \\
Rep. (\%) & 60.44 & 62.21 & 61.77  & 60.00  & 59.67 \\
Runtime (s)  & 0.74 & 0.79 & 0.72 & 0.69  & 0.74 \\
\bottomrule
\end{tabular}
}}
\end{center}
\caption{Ablation Study on Geometric Centroid Threshold $t_g$}
\label{table_ablation_geometric}
\end{table}
\vspace{-4pt}

\begin{table}[h]
\begin{center}
{\small{
\begin{tabular}{cccccc}
\toprule
$t_c$ & 0.1 & 0.2 & 0.3 & 0.4 & 0.5 \\
\midrule
\# Keypoint  & 616.20 & 565.67 & 524.31 & 495.34 & 477.46 \\
Rep. (\%) & 61.94 & 64.30 & 66.05 & 67.58 & 68.37 \\
Runtime (s)  & 0.70 & 0.74 & 0.68 & 0.71 & 0.66 \\
\bottomrule
\end{tabular}
}}
\end{center}
\caption{Ablation Study on Photometric Centroid Threshold $t_c$}
\label{table_ablation_color}
\end{table}
\vspace{-4pt}

\section{Conclusion}
In this work we propose the CEntroid Distance (CED) keypoint detector that can utilize both geometric and color information in colored point clouds for keypoint detection. 
We evaluate the proposed method against state-of-the-art handcrafted and learning-based keypoint detection methods on four synthetic and real-world datasets. Our method is demonstrated to be effective, stable and computationally efficient. We further demonstrate our method's potential to be used in applications such as colored point cloud registration.

We anticipate that the proposed CED detector can benefit systems that are capable of leveraging color information to improve performance, and that the centroid distance used herein can also serve as a stable measure to be used in other modalities. 
Future work also includes extension to multi-spectrum perception, and deployment of CED for autonomous robot navigation (e.g., \cite{kan2020online, kan2021task}) in mixed indoor and outdoor environments.

{\small
\bibliographystyle{ieee_fullname}
\bibliography{egbib}
}

\end{document}